\documentclass[journal,onecolumn]{IEEEtran}
\IEEEoverridecommandlockouts
\usepackage{cite}
\usepackage{amsmath,amssymb,amsfonts}
\usepackage{algorithmic}
\usepackage{graphicx}
\usepackage{textcomp}
\usepackage{xcolor}
\usepackage{caption}
\usepackage{subcaption}
\usepackage{enumitem}
\usepackage{comment}
\usepackage[inkscapelatex=false]{svg}
\def\BibTeX{{\rm B\kern-.05em{\sc i\kern-.025em b}\kern-.08em
    T\kern-.1667em\lower.7ex\hbox{E}\kern-.125emX}}
\begin{document}

\title{Human-Centered AI and Autonomy in Robotics: Insights from a Bibliometric Study\\
\thanks{This work has been funded by the PNRR - M4C2 - Investimento 1.3, Partenariato Esteso PE00000013 - ``FAIR - Future Artificial Intelligence Research" - Spoke 1 ``Human-centered AI" under the NextGeneration EU program, and by the Italian Ministry of University and Research (MUR) in the framework of the FoReLab and Cross-Lab projects (Departments of Excellence).}
}

\author{
    \IEEEauthorblockN{
        Simona Casini,
        Pietro Ducange,
        Francesco Marcelloni,
        Lorenzo Pollini
        }
    \IEEEauthorblockA{
        \begin{tabular}{cc}
            \begin{tabular}{@{}c@{}}
                    Department of Information Engineering, University of Pisa, Largo Lucio Lazzarino 1, 56122 Pisa, Italy\\
                    Email:
                    simona.casini@phd.unipi.it\\
                    \{pietro.ducange, francesco.marcelloni, lorenzo.pollini\}@unipi.it \\
                    \end{tabular}
        \end{tabular}
    }
}

\maketitle

\begin{abstract}

The development of autonomous robotic systems offers significant potential for performing complex tasks with precision and consistency. Recent advances in Artificial Intelligence (AI) have enabled more capable intelligent automation systems, addressing increasingly complex challenges. However, this progress raises questions about human roles in such systems. Human-Centered AI (HCAI) aims to balance human control and automation, ensuring performance enhancement while maintaining creativity, mastery, and responsibility. For real-world applications, autonomous robots must balance task performance with reliability, safety, and trustworthiness. Integrating HCAI principles enhances human-robot collaboration and ensures responsible operation. 

This paper presents a bibliometric analysis of intelligent autonomous robotic systems, utilizing SciMAT and VOSViewer to examine data from the Scopus database. The findings highlight academic trends, emerging topics, and AI’s role in self-adaptive robotic behaviour, with an emphasis on HCAI architecture. 
These insights are then projected onto the IBM MAPE-K architecture, with the goal of identifying how these research results map into actual robotic autonomous systems development efforts for real-world scenarios.
\end{abstract}


\section{Introduction} \label{introduction}

In recent decades, robotics has made significant advancements across various sectors, including aviation, transportation, marine, and agriculture.
According to the European strategy proposed by euRobotics in December 2024 \cite{eurobotics2024}, robotics is a complex integration of technologies that offers functional, economic, and societal benefits. Robots have evolved from performing isolated repetitive tasks to becoming integral components of infrastructures, such as farms, factories, transportation networks, and even homes, hospitals, and cities.
This progress is largely driven by two primary research areas: machine autonomy, which enables robots to operate and make decisions without human intervention, and machine automation, which focuses on designing robots to perform tasks with precision and consistency \cite{Wei2021}.
However, achieving full autonomy in Robotics and Autonomous Systems (RASs) presents challenges beyond just the design of robotic systems. These challenges include creating machines with human-level dexterity, proprioception, and physical awareness, as well as ensuring adaptability in dynamic environments. RASs must also redefine how machines interact with their surroundings and humans, especially given the growing demand for autonomous systems in complex, long-term scenarios. Autonomy here requires systems to adapt to external forces and disturbances over time, such as changes in weather or environmental conditions.

Artificial intelligence (AI) and its sub-disciplines provide powerful solutions that enhance traditional robotic approaches. The concept of \textit{intelligent automation} \cite{Tyagi2021} describes the integration of AI-driven decision-making and analytics into robotics, supporting autonomous processes. These systems, equipped with cognitive capabilities, can learn and adapt to evolving conditions. Key contributions to \textit{intelligent automation} research include Kam et al.'s review on research directions \cite{Kam2021}, and Wali et al.'s 2023 update \cite{Wali2023}, which incorporates mobile devices and graphical user interfaces into the scope of \textit{intelligent automation} platforms.
Despite these advancements, the integration of AI with traditional methods remains challenging, as highlighted by Moravec's paradox. This paradox points out that while machines may excel in complex tasks like playing chess, they struggle with tasks that are simple for humans, such as navigating a crowded room \cite{Oravec2022}. Thus, the challenge lies in identifying where and how AI and classical methodologies can be combined to optimize performance in areas like mapping, perception, and planning.

Over time, many researchers have approached RASs as intelligent automation systems, each exploring different aspects of the integration. A significant example is Kunze et al.'s study \cite{Kunze2018}, which provided an extensive survey of AI techniques as enablers of long-term autonomy in robotic systems. This study identified the challenges and future opportunities for AI in achieving long-term robotic autonomy. Borner et al. \cite{Borner2020} offered a literature review focused on the early 2000s, identifying three key domains of strategic interest: AI, robotics, and the Internet of Things (IoT). Kam et al. \cite{Kam2021b} further reviewed \textit{intelligent automation}, emphasizing that its adoption is driven by organizational strategy, business needs, and technological readiness.
While there are vertical reviews on specific AI applications within domains like marine or agricultural robotics, this study does not make assumptions about robot types or AI techniques, focusing instead on broader trends. A 2023 science mapping analysis by Hans \cite{Hans2023} examined the period 2016–2021, but it provided only a snapshot without a comprehensive view of the field’s evolution or future directions.

A common assumption in RAS research is that greater automation reduces user control. However, this overlooks the distinction between intervention and interaction. While automation can reduce the need for direct user intervention, it increases interaction by positioning the user as a supervisor. This distinction is crucial, raising questions such as: what is the role of the user in an autonomous system? How can RASs ensure reliability, safety, and trust in real-world applications?
Human-Centered AI (HCAI) principles \cite{Shneiderman2020} provide valuable insights here. HCAI emphasizes that systems should be comprehensible, predictable, and controllable to enhance user confidence. IBM's AI fundamentals \cite{IBMweb} similarly highlight the importance of human-machine collaboration, where machines augment human abilities rather than replace them. Examples of HCAI in robotics, such as the works of Doncieux et al. \cite{Doncieux2022} and He et al. \cite{He2022}, explore this synergy further.

This study offers a comprehensive overview of the evolution of AI techniques in robotics, focusing on RAS as \textit{intelligent automation} systems. It traces historical advancements, highlights future challenges, and emphasizes the importance of Human-Centered design principles to ensure that robotic systems align with user needs. These principles aim to ensure that robotic systems align with user needs by prioritizing trustworthiness, safety, and meaningful human-robot interactions.
The paper is structured as follows: Section \ref{sec: methodology} introduces the methodology detailing the research questions and the science mapping tools used. Section \ref{sec: descriptive_analysis} presents a descriptive analysis, followed by a deeper exploration of science maps in Section \ref{sec: map_analysis}. Section \ref{sec: results} summarizes the findings addressing both past achievements and future interests. In Section \ref{sec: MAPE-K} we propose the integration of these insights into a state-of-the-art architecture. Finally, Section \ref{sec: conclusion} draws conclusions.  

\section{Methodology} \label{sec: methodology}

\subsection{Research question and filters} \label{sec: question}

As discussed in the previous section, RASs have benefited from the integration of AI techniques, enabling several innovations. Thus, understanding the evolution of these advancements means understanding the historical development, current trends, and future directions of AI-driven robotics.
To address these aspects, this study leverages a comprehensive gathering and analysis of relevant scientific papers using the Scopus database, since it is considered a de-facto standard in document collection.

The main research question guiding the authors is:

\textbf{Q: How do AI algorithms contribute to the self-adaptive behaviour of a robot in a fully autonomous real-world scenario?}

In addition to this primary question, the study aims to explore two specific sub-questions:

\textbf{Q1: What is the historical evolution of AI contributions in autonomous systems? Are there evident trends or milestones that can be identified?}

\textbf{Q2: Based on the progression over the years and emerging topics or keywords, what are the potential academic directions for the near future?}

The relevant literature was retrieved using a query that includes both general keywords, such as \textit{artificial intelligence}, \textit{machine learning}, \textit{autonomous robot*} and \textit{data-driven}, as well as more specific terms like \textit{neural network}. Moreover, we limited the language to English and the publication period to 2000-2024 range. The result is a set of 2,564 articles.

\subsection{Science mapping tools}\label{subsec: tools}

\begin{figure}
     \centering
     \begin{subfigure}[b]{0.48\textwidth}
         \centering
         \includegraphics[width=5 cm]{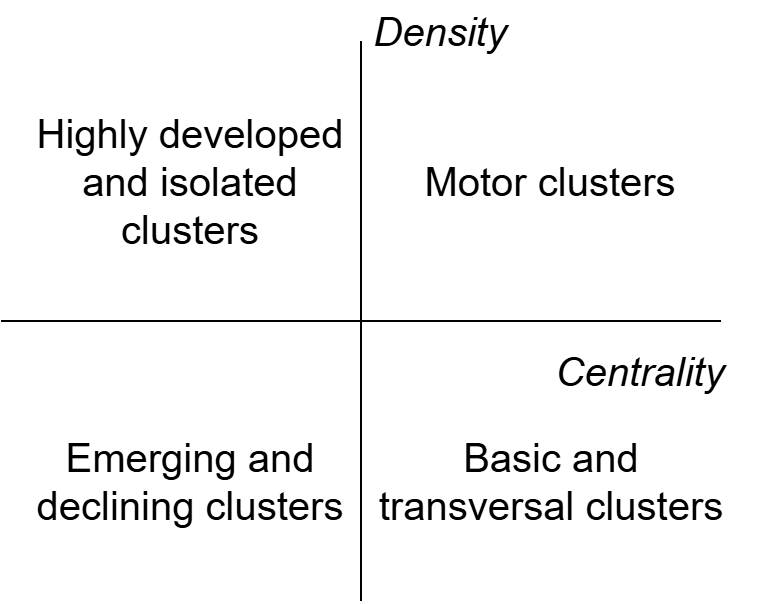}
         \caption{strategic diagram}
         \label{fig:strategic_diagram}
     \end{subfigure}
     \hfill
     \begin{subfigure}[b]{0.48\textwidth}
         \centering
         \includegraphics[width=4 cm]{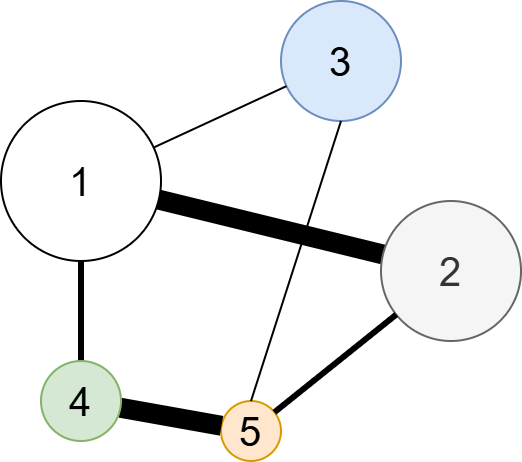}
         \caption{a generic example of cluster network}
         \label{fig:cluster_network}
     \end{subfigure}
     \hfill
        \caption{SciMAT Maps description - Details in \cite{Cobo2011}.}
        \label{fig:three graphs}
    \vspace{-15pt}
\end{figure}

To comprehend the evolving field of AI, a systematic approach is crucial for synthesizing existing evidence. The vast volume of literature poses risks such as biases and knowledge gaps, which can compromise the reliability of outcomes. Bibliometric analysis, or science mapping, offers an objective alternative. This method employs tools to create maps that reveal leading research themes, capturing the knowledge structure of a domain and providing insights into its connections [17], [18]. The science mapping process typically involves key steps: data collection, preprocessing, network extraction, mapping, metric definition, and visualization [19]. The results then require thorough analysis. A review by [20] identifies nine widely used science mapping tools, each suited to different stages of the process.

This study employs two tools: VOSviewer and SciMAT. VOSviewer \cite{VanEck2010}, developed in 2019, excels in visualizing large datasets through \textbf{distance/proximity-based maps}, where item proximity reflects relationship strength. Its minimal adjustments make it efficient for bibliometric data visualization.
SciMAT \cite{Cobo2012}, developed in 2012, handles flexible knowledge analysis and customizable metrics, providing a powerful tool for trend identification and understanding themes' influence.
While VOSviewer visualizations are intuitive, SciMAT maps require additional interpretation.

Three types of SciMAT graphs, namely the strategic diagram, the cluster network and the evolution map, are considered in this paper. 
One widely used methodological approach for identifying research themes within a scientific domain is co-word analysis, which is applied to the set of all published documents during a specified period. This method systematically examines the relationships between keywords to uncover patterns of co-occurrence that highlight interconnected research themes. Through this process, clusters of keywords are formed, representing networks of documents that address closely related topics or widely studied problems in the field.
To measure the strength of connections between keywords, the equivalence index is employed, quantifying how often specific keywords appear together across publications \cite{Cobo2011}. These connections form a co-word network, which is then partitioned into distinct clusters, each representing a specific research theme. Within each cluster, keywords are strongly linked due to their frequent co-occurrence, indicating a focused area of inquiry.
The resulting themes are visualized as spheres in thematic maps, with each sphere corresponding to a cluster of closely related keywords. Clusters provide a detailed representation of the intellectual structure of the field, capturing the central research interests and trends.

The relative contribution of each research theme to the overall field is evaluated both quantitatively and qualitatively; this assessment highlights the most prominent, productive, and high-impact subfields using bibliometric indicators, such as the number of published documents and average citations per theme. Larger values of these indicators are visualized on the map by larger volumes of the themes' spheres.

The \textbf{strategic diagram} (Figure \ref{fig:strategic_diagram}) positions clusters in a two-dimensional space using density and centrality metrics. Density measures the internal strength of the network. Centrality measures the degree of interaction of a network with other networks. 
The 'low' and 'high' values of these two metrics divide the two-dimensional space into four quadrants: 
\begin{itemize}[leftmargin=*]
    \item motor clusters (upper-right): high density and high centrality, central to the field; 
    \item highly developed and isolated clusters (upper-left): high density and low centrality, niche but robust; 
    \item emerging and declining clusters (lower-right): low density and high centrality, either growing in relevance or losing significance;
    \item basic and transversal clusters (lower-left): low density and low centrality, foundational or underdeveloped. 
\end{itemize} 

The \textbf{cluster network} (Figure \ref{fig:cluster_network}) complements the strategic diagram by visualizing relationships within the main cluster and its sub-network, highlighting structural dynamics. Key elements include: the main cluster, which represents the central theme, the sub-network items, such as related topics or authors, and edge weights, which indicate the strength of relationships - thicker edges denote stronger connections.

In the end, the \textbf{evolution map} (an example is shown  in Fig. \ref{fig: history}) analyses the temporal development and connections between research themes across different time periods. It helps to identify the evolution of clusters, and keywords by tracking their presence, relationships, and transitions over time. Edges between themes indicate connections similar to those in a cluster network. A solid line signifies that the linked themes share keywords (either the same label or through label inclusion), while a dotted line indicates shared elements that do not correspond to the theme's name. The thickness of the edges reflects the inclusion index, with greater width denoting a higher number of shared elements in the sets’ intersection. Finally, vertical lines demarcate different periods. 

Together, the three SciMat graphs offer a detailed view of thematic relevance, cohesion, and interconnections within the research domain in the given period. Further details can be found in \cite{Cobo2011}.
In summary, VOSviewer was used to generate proximity maps emphasizing clusters and their interconnections, while SciMAT was used to perform two main analyses, leveraging data from 2000 to 2024.
The first one generates one single historical diagram, divided into three areas, each corresponding to a distinct decade. The third "decade" (2020–2024) spans only five years but includes the highest number of publications, reflecting the recent surge in research activity. This approach enables the examination of the temporal evolution of key research clusters/themes.
The second analysis covers instead the entire period (2000–2024), providing a comprehensive overview of 25 years of research. This global perspective highlights emerging clusters and identifies those that have maintained consistent relevance throughout the time-frame.

The following sections provide both a general dataset descriptive analysis and detailed comments based on the integrated interpretation of the obtained maps.

\section{Descriptive analysis} \label{sec: descriptive_analysis}

This section provides a preliminary qualitative overview of the data obtained from the query on Scopus, focusing on publication trends, geographical distribution, document types, and keyword analysis.

\begin{figure}
    \centering
    \begin{subfigure}[b]{0.32\textwidth}
        \includegraphics[width=1\textwidth]{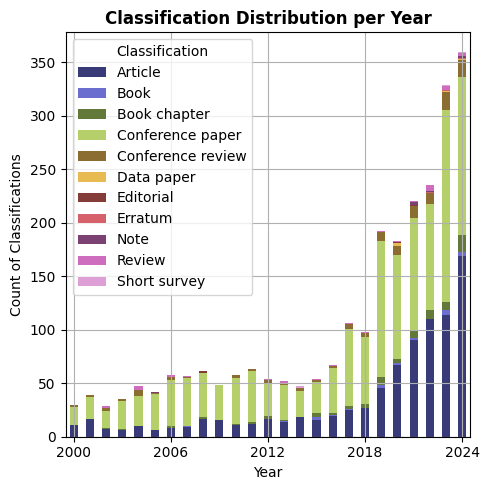}
        \caption{Documents distribution per year}
        \label{fig:papers per year}
     \end{subfigure}
    \begin{subfigure}[b]{0.33\textwidth}
        \includegraphics[width=1\textwidth]{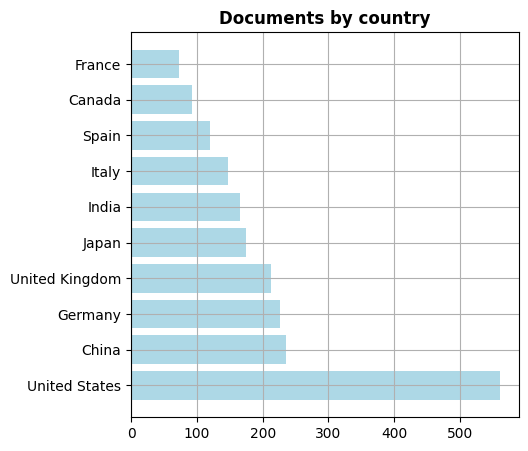}
        \caption{First 10 most productive countries}
        \label{fig:papers per country}
     \end{subfigure}
    \begin{subfigure}[b]{0.32\textwidth}
        \includegraphics[width=1\textwidth]{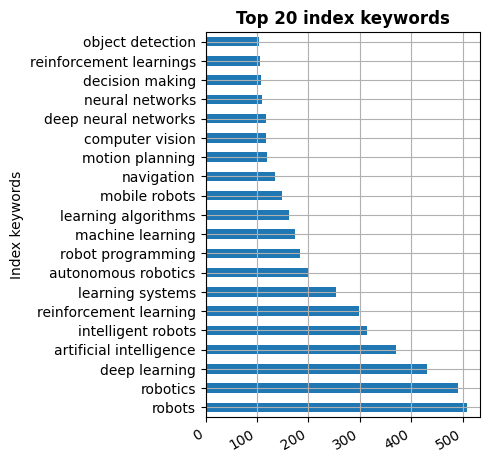}
        \caption{First 20 most recurrent keywords}
        \label{fig:index keywords}
     \end{subfigure}    
\end{figure}

\textbf{Publication trends.} Figure \ref{fig:papers per year} highlights the steady growth in publications on this topic since 2000, with a significant increase starting from 2017. This increase is not solely attributable to advancements in robotics but is closely tied to the rapid growth of AI research. As noted in \cite{Rane2024}, \cite{Baruffaldi2020}, and \cite{Aggarwal2022}, this trend can be explained by a combination of technological, economic, and social factors. These studies also provide insights into the potential future directions and societal impacts of AI applications.

\textbf{Geographical distribution.} Figure \ref{fig:papers per country} lists the 10 most productive countries. The USA leads with twice as many publications as China, the second-ranking country. This dominance underscores the substantial investment in AI and robotics research within the USA, driven by robust academic and industrial initiatives.

\textbf{Document types.} The majority of contributions are conference papers, followed by book chapters (figure \ref{fig:papers per year}). This reflects the dynamic nature of the field, where conference presentations are preferred for disseminating cutting-edge research, often preceding journal publication.

\textbf{Keyword analysis.} Figure \ref{fig:index keywords} shows the 20 most frequently used Index (i.e. Scopus assigned) keywords. These provide an initial glimpse into the topics covered in the literature. General terms such as \textit{robotics}, \textit{artificial intelligence}, and \textit{learning systems} dominate, reflecting the broad scope of the field. However, specific techniques like \textit{reinforcement learning} and \textit{deep learning}, as well as applications such as \textit{navigation}, \textit{computer vision}, and \textit{motion planning}, hint at the methodologies and challenges addressed by researchers. 

\section{Historical analysis} \label{sec: map_analysis}

\begin{figure*}
    \centering
    \includegraphics[width=0.9\textwidth]{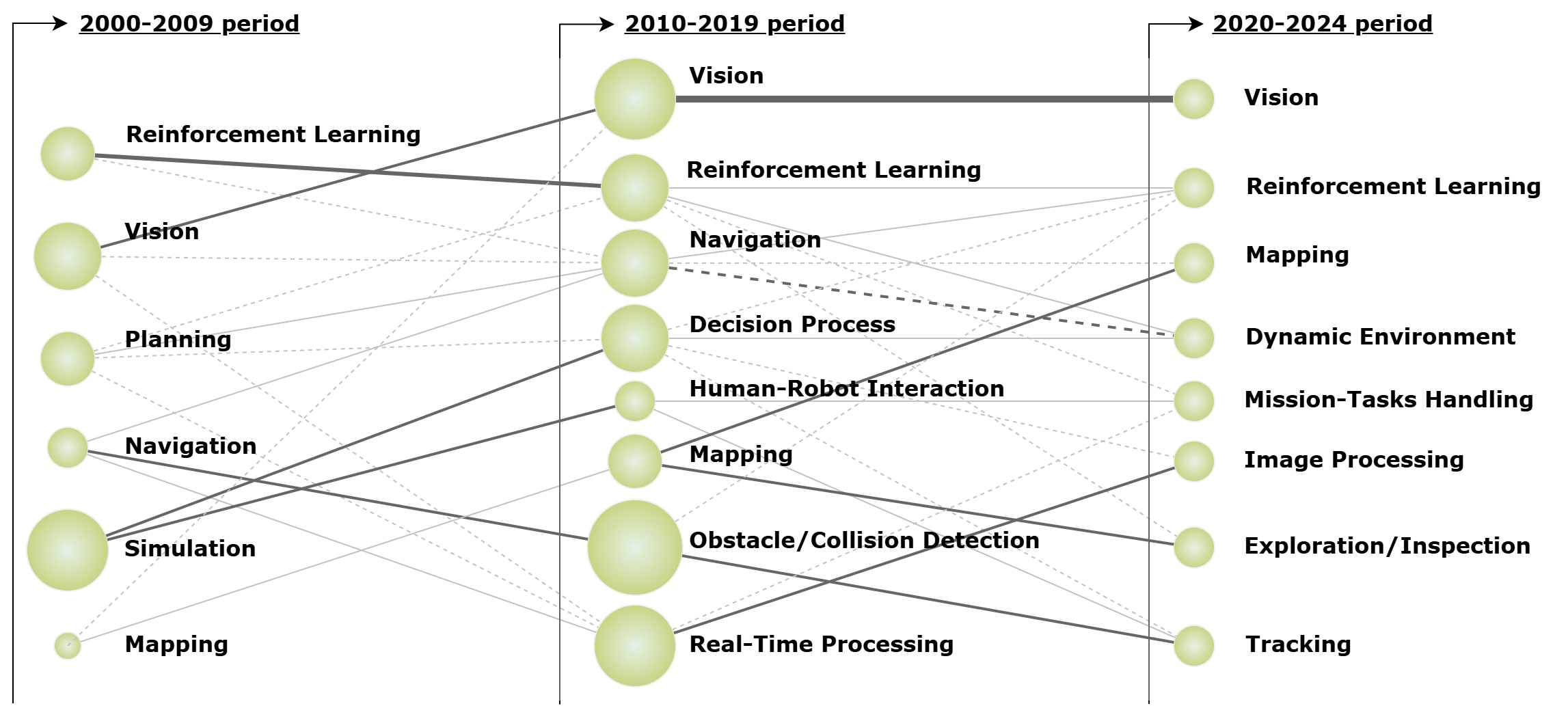}
    \caption{Evolution map - The average number of citations is used as the bibliometric indicator (see subsection \ref{subsec: tools}). Consider, for example \textit{Mapping} cluster. Over the three decades, the cluster has progressively evolved, particularly in the most recent period, into a more complex task domain, such as exploration and inspection, following its convergence with the \textit{reinforcement learning} cluster.}
    \label{fig: history}
\end{figure*}

Starting with individual decades and then examining the entire period as a whole, this section analyses the maps obtained from SciMAT and VOSviewer.

Figure \ref{fig: history} shows the evolution map. It paints a visual narrative of how research in Human-Centered AI and autonomy in robotics has developed over time, considering both continuity and transformation in the field. The diagram shows the three consequent periods considered and offers a look into the dynamic interplay of themes, showcasing their rise, persistence, and interconnection. 
At the core of this narrative lies the enduring significance of \textit{reinforcement learning} and \textit{vision}, which serve as foundational pillars throughout all periods. These themes maintain their centrality, with strong connections to other areas, underscoring their critical role in advancing robotics and AI. Over time, their influence expands, driving progress in both theoretical and applied domains.
As the timeline progresses, the emergence of new themes reflects a shift toward addressing practical challenges. For instance, \textit{Human-Robot interaction} gains prominence in the middle period, highlighting a growing focus on making robotics more intuitive and user-friendly, a key tenet of HCAI. This shift emphasizes the importance of designing systems that prioritize human needs, accessibility, and collaboration. Similarly, the rise of \textit{obstacle/collision detection} and \textit{real-time processing} in later periods signals an increased emphasis on safety, responsiveness, awareness and real-world applicability, aligning with HCAI principles that advocate for trustworthy and reliable AI systems.
One of the most compelling aspects of this map is its depiction of an evolving research ecosystem. Foundational themes like \textit{mapping}, \textit{navigation}, and \textit{simulation} interact with emerging topics such as \textit{decision process} and \textit{dynamic environment}, reflecting a shift toward systems that are not only autonomous but also capable of adaptive and intelligent decision-making. From an HCAI perspective, this evolution underscores the need for AI systems that are not only technically robust but also ethically aligned and capable of enhancing human decision-making rather than replacing it.
This first analysis well underscores the dynamic nature of the field, illustrating the emergence of specialized topics alongside the continued evolution of foundational themes. It also highlights the growing alignment of AI research with HCAI principles, emphasizing the importance of human-centric design, safety, and ethical considerations in shaping the future of intelligent systems.

\begin{figure}
    \centering
    \includegraphics[width=0.45\textwidth]{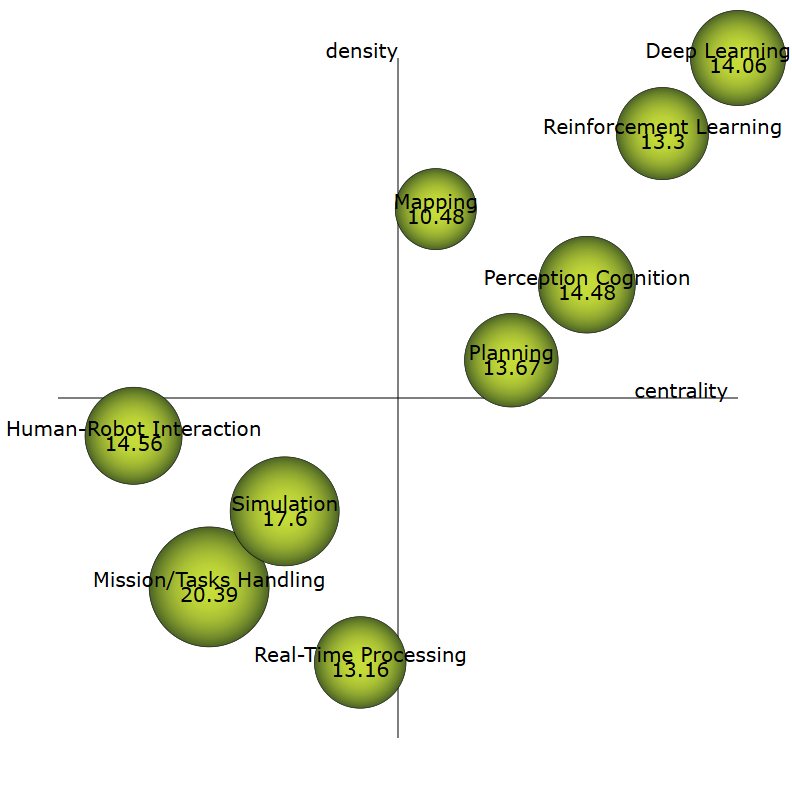}
    \caption{Strategic diagram from 2000 to 2024. The average number of citations is used as the bibliometric indicator (see subsection II-B)}
    \label{fig:0024}
\end{figure}

Considering the entire period from 2000 to 2024, several additional insights emerge, as shown in Figure \ref{fig:0024}. The strategic map shows that only motor and emergent cluster quadrants are populated through centrality and density index, confirming what was discussed previously: no isolated clusters are present.
In motor clusters, \textit{deep learning} and \textit{reinforcement learning} keywords show the most significant impact. Once again the RAS awareness emerges as a priority appearing in \textit{perception and cognition}, \textit{mapping} and \textit{planning} clusters.
The importance of explainability, which includes providing informative feedback, progress indicators, and detailed completion reports (i.e. HCAI Prometheus Principles \cite{Shneiderman2020}) is once again confirmed. \textit{mission handling}, \textit{real-time processing} and \textit{Human-Robot interaction} are revealed not only as themes (fig. \ref{fig: history}) but also as highly developed and cited individual clusters.

\begin{figure}
    \centering
    \begin{subfigure}[b]{0.32\textwidth}
        \centering
        \includegraphics[width=1\textwidth]{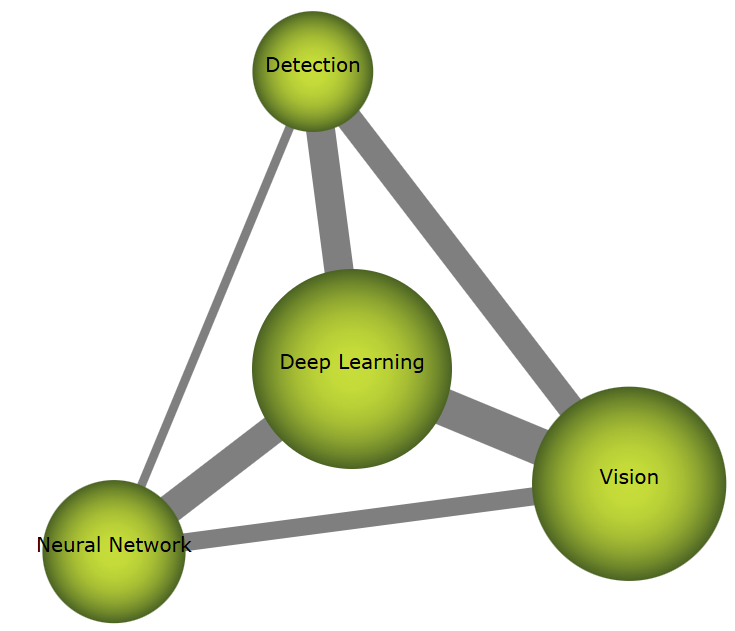}
        \caption{Deep learning cluster network}
        \label{fig:deep learning cluster network}
    \end{subfigure}
    \begin{subfigure}[b]{0.32\textwidth}
        \centering
        \includegraphics[width=1\textwidth]{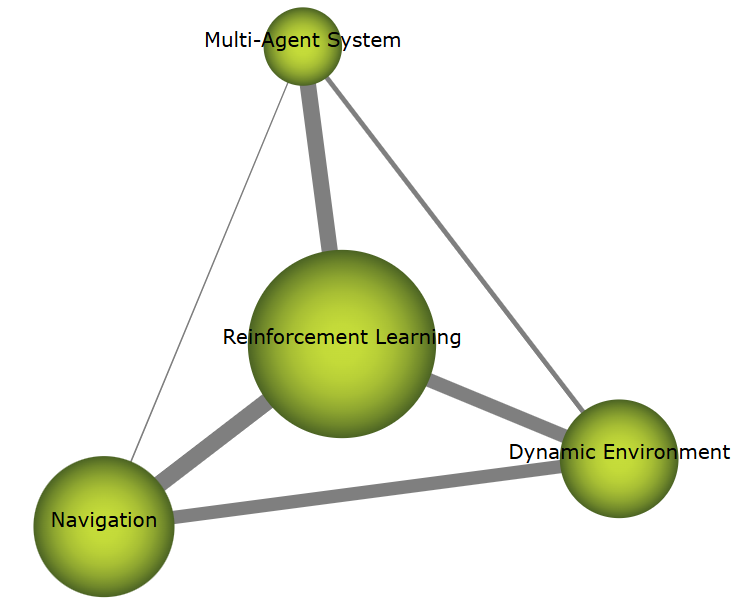}
        \caption{RL cluster network}
        \label{fig:reinforcement learning cluster network}
    \end{subfigure}
    \begin{subfigure}[b]{0.32\textwidth}
        \centering
        \includegraphics[width=1\textwidth]{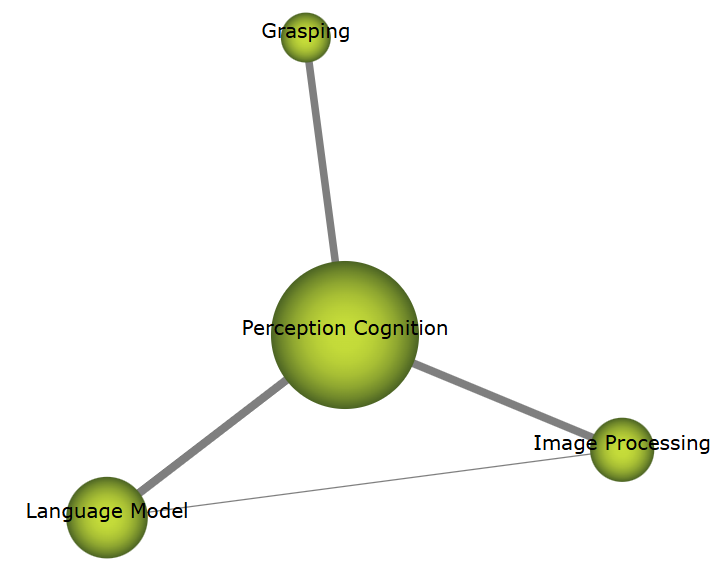}
        \caption{Perception/Cognition cluster network}
        \label{fig:artificial intelligence cluster network}
    \end{subfigure}
\end{figure}

A deeper cluster understanding can be achieved by examining the network diagrams of the three main motor clusters (selected based on centrality/density index) to visualize their related areas of interest.
In the \textit{deep learning} cluster, which includes more than 800 documents, Figure \ref{fig:deep learning cluster network} reveals a strong relationship between deep learning methods (particularly neural networks) and \textit{vision} applications, primarily for \textit{detection} tasks (objects, obstacles, people, and other entities in the environment).
The \textit{reinforcement learning} cluster, with more than 500 documents, identifies two primary applications (Figure \ref{fig:reinforcement learning cluster network}): \textit{navigation} tasks, encompassing control methods during execution, particularly in \textit{dynamic environments}, and high-level task management. The latter involves applying reinforcement learning techniques for strategy policy both for individual robots and within \textit{multi-agent systems}.
The \textit{perception cognition} cluster network, even if seems directly related to vision processes and thus neural network algorithms, reveals a connection with another interesting theme: \textit{language model}. Increasing capabilities of natural language models, open possibilities not only in the communication area but also in RAS awareness. A great example can be found in \cite{Huang2022} where an LLM chatbot is proposed to perform a pipeline of a \textit{grasp} task. Considering this case study, HCAI transparency in decision-making processes not only built trust, but also empowers users to understand, validate, and collaborate with AI-driven robotic systems effectively — an essential requirement at every stage of RAS development, from design to deployment.

\begin{figure}
    \centering
    \includegraphics[width=0.6\textwidth]{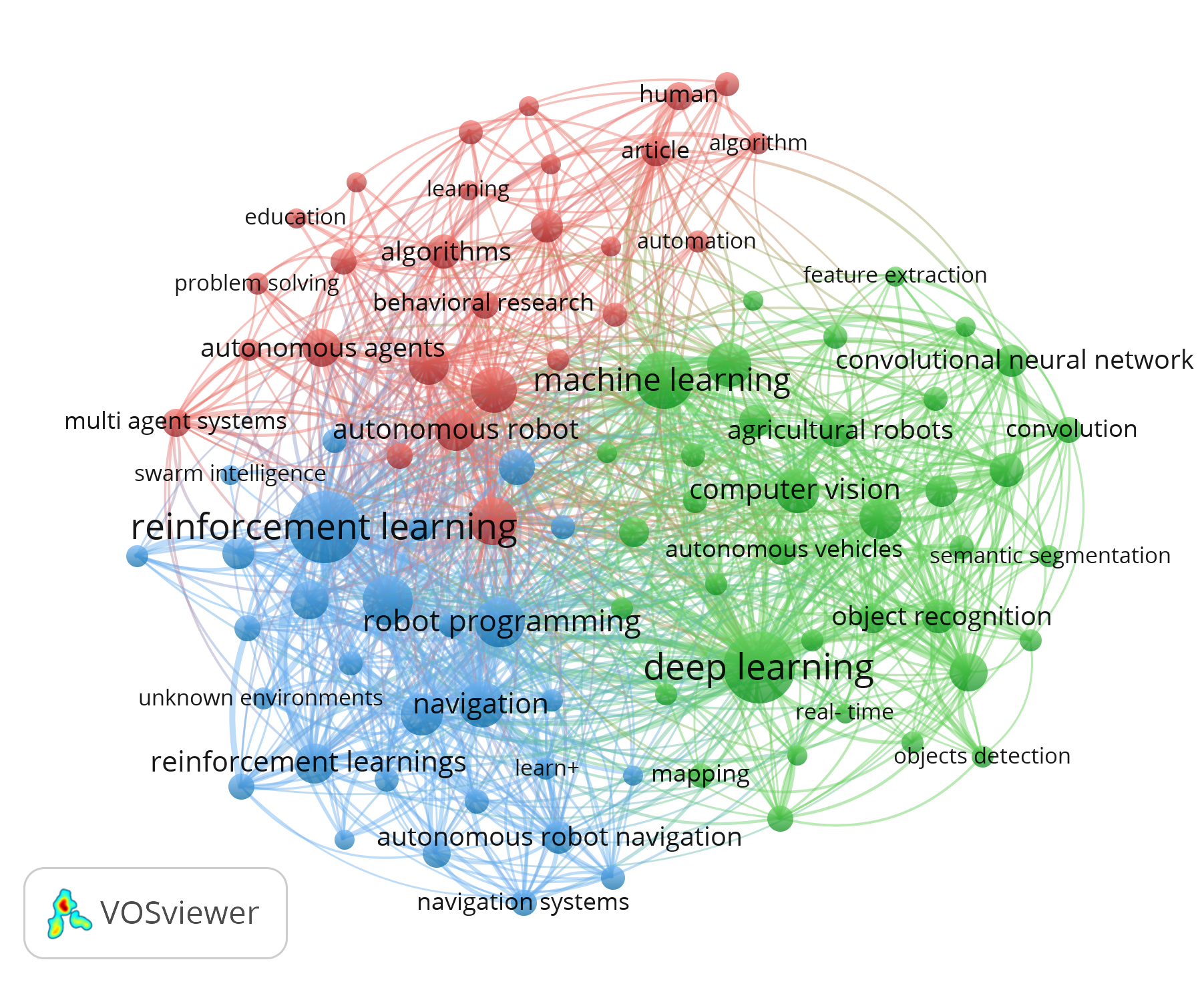}
    \caption{VOSViewer network visualization}
    \label{fig:network VOSViewer}
\end{figure}

As noted earlier, SciMAT was not the sole bibliometric tool employed in this analysis. Shifting focus to the results from VOSviewer, Figure \ref{fig:network VOSViewer} illustrates three primary clusters identified through the \textit{distance-based} calculation method.

\begin{itemize}[leftmargin=*]
    \item The red cluster represents high-level control tasks and knowledge, containing keywords such as \textit{interaction}, \textit{human}, \textit{decision}, as well as general terms like \textit{problem solving} and \textit{behavioural research};
    \item the blue cluster includes low-level control and environment-related features, such as \textit{navigation}, \textit{reinforcement learning}, \textit{unknown environment}, and \textit{obstacle}. This cluster is closely associated with the planning and execution phases;
    \item the green cluster corresponds to monitoring tasks, focusing on sensor-related processes. Keywords such as \textit{vision}, \textit{object}, \textit{detection}, and \textit{image} are prominent, indicating that the localization phase is central to this cluster. \textit{Neural networks} appear as the main approach to solve these functions.
\end{itemize}
Examining the mutual relationships between keywords reveals that there is no distinct separation between clusters. All topics are interconnected, with a maximum of two degrees of separation. This suggests that the identified topics are part of an integrated system, where no single area dominates over the others.
The keywords, macro-clusters, and intrinsic relationships identified through VOSviewer align well with the network analysis from SciMAT. In particular, the technique/task binomials correspond between the two tools: for example, \textit{reinforcement learning} is strongly linked to navigation and planning, while \textit{deep learning} is associated with vision and localization.

\section{Results summary and Future Directions} \label{sec: results}

The following items summarize the results from the descriptive and mapping analyses:

\begin{itemize}[leftmargin=*]
    \item AI in robotics is a transversal concept, applicable to various aspects of robot behaviour. Interest in this field has grown steadily between 2000 and 2024, with a significant increase observed after 2017;
    \item over the decades, particularly since 2020, the study and application of advanced AI techniques have enabled RASs to address increasingly complex challenges, including those involving unknown and dynamic environments;
    \item emerging areas of interest include \textit{perception} and \textit{cognition}, which focus on robot awareness, often in conjunction with \textit{navigation}. This also encompasses Simultaneous Localization and Mapping (SLAM) techniques;
    \item AI has been seen to enter crosswise into every part of the implementation process of the RAS. HCAI thus needs to become a paradigm shift from the development to the deployment of AI technologies and becomes crucial from lower-level implementation (i.e. simulation) to high-level communication and decision-making. 
\end{itemize}
The predominant AI techniques applied are:
\begin{itemize}[leftmargin=*]
    \item deep learning and neural network, primarily used in vision tasks for object detection and recognition. These applications include classification problems related to localization, motion planning, and path planning;
    \item reinforcement learning, employed in complex tasks involving high-level robot behaviour, interaction, and decision-making layers. It is also applied to navigation problems in unknown or dynamic environments, such as map generation and interpretation, localization, and path planning.
\end{itemize}

\begin{figure}
    \centering
    \includegraphics[width=0.45\textwidth]{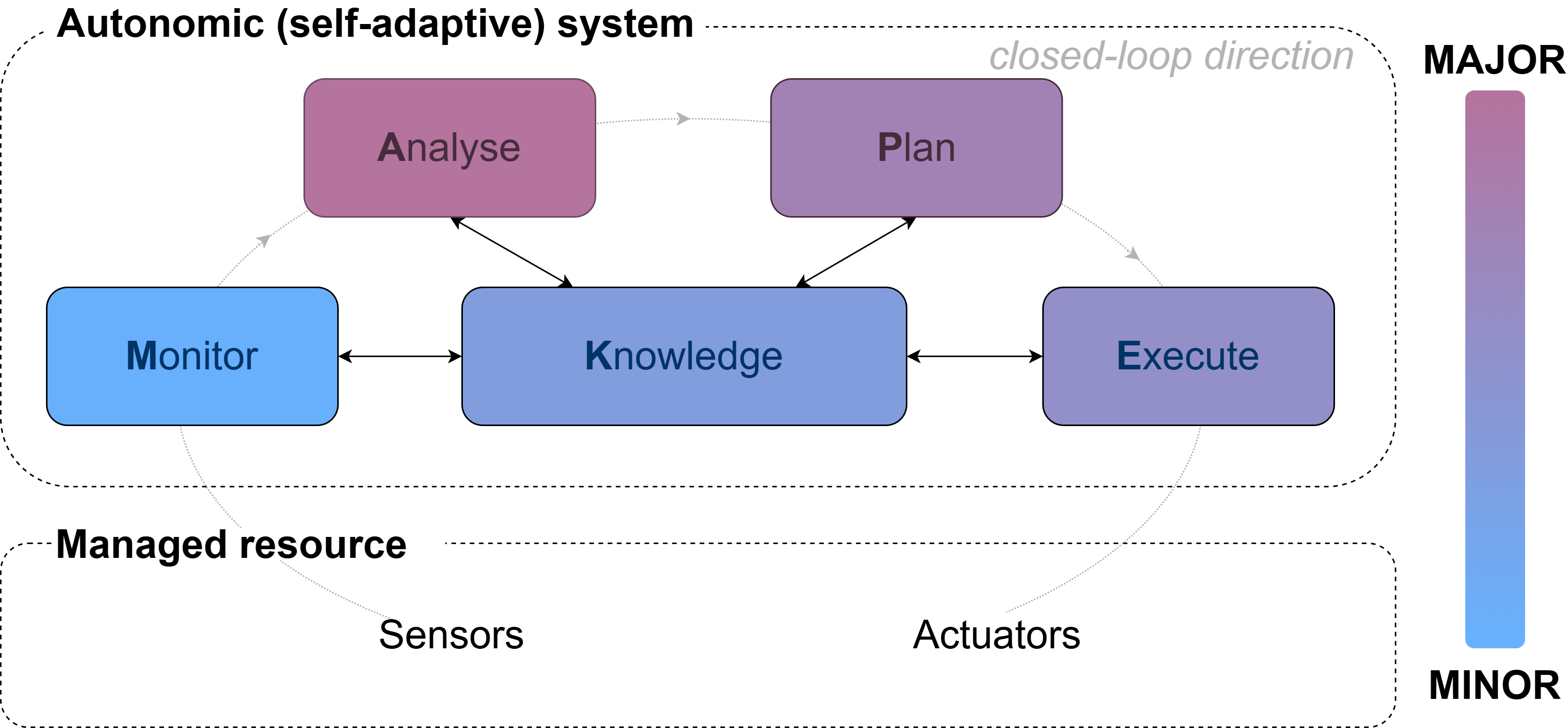}
    \caption{MAPE-K framework with colours reflecting historical research interests based on bibliometric analysis.}
    \label{fig:MAPE-K summary}
\end{figure}
As regards future directions, what seems to emerge from themes evolution but also from clusters centrality, can be summarized as follows:
        \begin{itemize}[leftmargin=*]
    \item \textbf{uncertainty management:} addressing perturbances in RAS, particularly in unknown or dynamic environments. These two scenarios may coincide when the environmental model is initially unknown;
    \item \textbf{mission and task handling in real-world scenarios:} in complex environments, a robot's mission often needs to be divided into multiple sub-tasks or objectives. These must be managed and completed with the desired performance metrics. This area is particularly intriguing as it allows for exploring controversial approaches, such as counterfactual thinking \cite{Neal2017}, an intrinsic aspect of human behaviour that remains difficult to replicate in robotic systems. Additionally, the HCAI paradigm aligns with this challenge, as it emphasizes providing users with a clear understanding of the machine's state, maintaining control where necessary, and merging human expertise with robotic competencies to enhance overall performance;
    \item \textbf{Human-robot and robot-robot} (multi-agent systems) \textbf{interaction} for shared autonomy tasks: this area further aligns with the HCAI paradigm, reinforcing the need for effective collaboration and interaction to achieve shared goals. It also ties directly to the previous challenge, creating a cohesive framework for future research and implementation.
\end{itemize}
The future objectives identified are also closely related to the key technical challenges outlined in the euRobotics strategy. These include \textit{harsh environment operations}, \textit{endurance features} (extended periods operations in complex environments without human intervention and with high dependability), \textit{human interaction}, and \textit{knowledge integration} (understood as the semantic understanding of the environment and skill exchange).
This continuity between the findings of the literature review and the priorities highlighted by the European committee is particularly significant. It underscores real needs that have already been acknowledged by parts of the research community. This convergence also paves the way for potential collaborations between research institutions and industrial partners, fostering advancements in the field.

\begin{figure}
    \centering
    \includegraphics[width=0.45\textwidth]{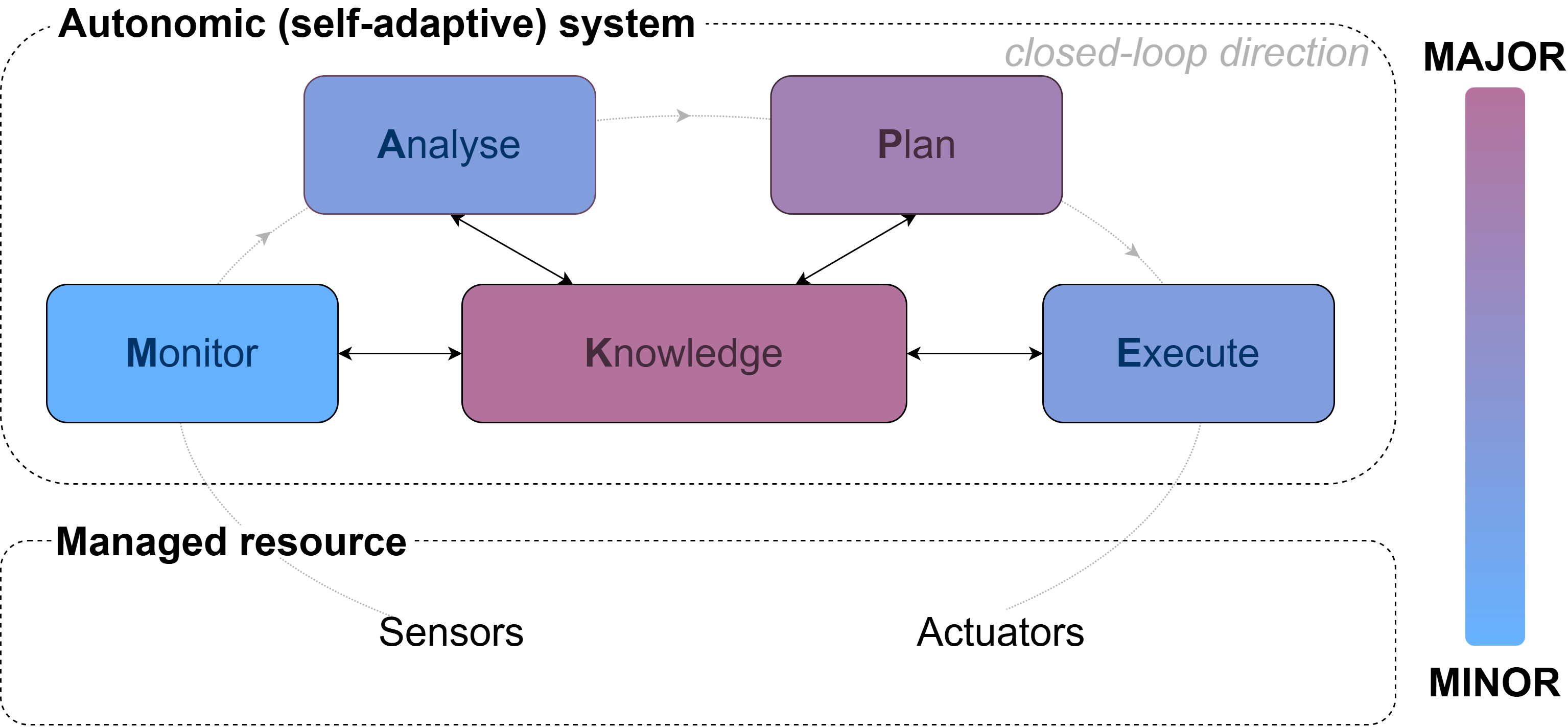}
    \caption{MAPE-K framework with colours reflecting potential future research directions from trend analysis.}
    \label{fig:MAPE-K summary direction}
\end{figure}

Insights derived from the science mapping can be interpreted and applied at different levels of abstraction. As an additional contribution to this study, a transposition of the identified findings into a classic state-of-the-art autonomous system architecture is proposed. This representation visualizes the main areas of scientific contribution and potential future directions within a closed feedback loop.

\section{MAPE-K Architecture for HCAI-based Autonomous Robotic Systems} \label{sec: MAPE-K}

An architecture for robot autonomy should define layers of abstraction that describe the components that let autonomous systems operate, adapt, and optimize in real-world environments. Key principles should include self-monitoring, self-diagnosis, and self-optimization, typically implemented through closed-loop feedback.
A leading model for autonomous systems is IBM's MAPE-K framework (Monitor, Analyse, Plan, Execute and Knowledge) \cite{White2004}. This framework supports self-managing systems, enabling adaptation to internal and external changes (see Figures \ref{fig:MAPE-K summary} and \ref{fig:MAPE-K summary direction}). While operating at a high level of abstraction, MAPE-K is particularly suited for robotics, aligning with fundamental robotic functions like Sense, Plan, and Act.
\begin{comment}
\begin{figure}
    \centering
    \includegraphics[width=0.45\textwidth]{figures/MAPE-K_general.png}
    \caption{IBM's MAPE-K framework.}
    \label{fig:MAPE-K}
\end{figure}
\end{comment}
In robotics, MAPE-K can be extended to address both high-level mission planning and low-level control. From an HCAI outlook, the \textit{Plan} and \textit{Knowledge} components are crucial. The \textit{Plan} component supports mission handling by breaking down complex tasks and ensuring transparency for human users, while the \textit{Knowledge} component enables users to express and refine their intent, fostering collaboration and trust.

The 2564 papers were mapped into the MAPE-K framework by analysing them and classifying the papers by relevance into the 5 main subfunctions: Monitor, Analyse, Plan, Execute and Knowledge. 
Integrating bibliometric insights into MAPE-K reveals key areas of scientific contribution and future focus. The \textit{Analyse} block, covering robot awareness, localization, and detection, and the \textit{Plan} block, encompassing navigation and task planning, have been central in past research. The \textit{Execute} block remains vital, focusing on control processes and translating variables into robot actions.

Future challenges emphasize the \textit{Knowledge} block, which integrates the robot's self-awareness and environmental understanding while facilitating communication with humans and other robots in multi-agent systems. These capabilities are further enhanced by significant advancements in Large Language Models, currently being tested to enable intuitive human-robot interaction. Mission handling, within the \textit{Knowledge} and \textit{Plan} blocks, allows for task decomposition, improving traceability and adaptability. Uncertainty management spans across the \textit{Analyse}, \textit{Knowledge}, and \textit{Execute} blocks, addressing sensor fusion errors, model inaccuracies, and external disturbances.
Figures~\ref{fig:MAPE-K summary} and~\ref{fig:MAPE-K summary direction} visualize the bibliometric results and future directions within the MAPE-K framework, with colour gradients from blue to purple indicating increasing interest in past and future research.

In conclusion, the MAPE-K framework provides a robust foundation for designing trustworthy and reliable robotic systems. Its integration with HCAI principles ensures transparency, adaptability, and collaboration, paving the way for advancements in autonomous robotics and their application in real-world scenarios.

\section{Conclusions} \label{sec: conclusion}

In this paper, a bibliometric analysis of the application of artificial intelligence (AI) in autonomous robotic systems, covering the period from 2000 to 2024 has been presented. Using SciMAT and VOSviewer tools, the study has outlined the field's evolution, key techniques, and emerging trends, highlighting AI's critical role as a transversal enabler in robotics.
The steady growth of publications, particularly after 2017, reflects increasing interest in leveraging AI to address complex tasks in dynamic and uncertain environments. Techniques such as deep learning and reinforcement learning have been pivotal in enhancing vision, navigation, and decision-making capabilities, cementing their relevance in both foundational research and applications.
Future advancements must address uncertainties in dynamic environments, enhance mission handling through sub-task management, and improve human-robot interaction in shared autonomy tasks. 
HCAI-based vision emerges thus as a transversal theme, emphasizing transparency, trust and clear communication. 
These principles, integrated with the MAPE-K framework, provide a robust foundation for self-monitoring, adaptability, and accountability in robotic systems. The \textit{Plan} and \textit{Knowledge} components, in particular, supporting mission decomposition, adaptability, and user collaboration, ensure systems remain trustworthy while achieving high performance.

In summary, integrating advanced AI techniques with frameworks like MAPE-K and HCAI offers immense potential for addressing complex challenges. A continued focus on human-centric principles will ensure that these systems remain ethical, trustworthy, and impactful, contributing significantly to advancements in robotics and their integration into daily life.


\begin{thebibliography}{99}
\bibitem{eurobotics2024}
https://eu-robotics.net/strategy/

\bibitem{Wei2021}
Xu, Wei. (2021). From Automation to Autonomy and Autonomous Vehicles: Challenges and Opportunities for Human-Computer Interaction. interactions. 28. 10.1145/3434580.

\bibitem{Tyagi2021}
Tyagi, Amit Kumar, et al. "Intelligent automation systems at the core of industry 4.0." International conference on intelligent systems design and applications. Cham: Springer International Publishing, 2020.

\bibitem{Kam2021}
Ng, Kam KH, et al. "A systematic literature review on intelligent automation: Aligning concepts from theory, practice, and future perspectives." Advanced Engineering Informatics 47 (2021): 101246.

\bibitem{Wali2023}
Wali, A.; Mahamad, S.; Sulaiman, S. Task Automation Intelligent Agents: A Review. Future Internet 2023, 15, 196.

\bibitem{Oravec2022}
Oravec, J.A., "Good Robot, Bad Robot: Dark and Creepy Sides of Robotics, Autonomous Vehicles, and AI", Springer International Publishing, Series Social and Cultural Studies of Robots and AI, 2022, ISBN 9783031140136

\bibitem{Kunze2018}
L. Kunze, N. Hawes, T. Duckett, M. Hanheide and T. Krajník, "Artificial Intelligence for Long-Term Robot Autonomy: A Survey," in IEEE Robotics and Automation Letters, vol. 3, no. 4, pp. 4023-4030, Oct. 2018

\bibitem{Borner2020}
Börner, K., Scrivner, O., Cross, L. E., Gallant, M., Ma, S., Martin, A. S., ... \& Dilger, J. M. (2020). Mapping the co-evolution of artificial intelligence, robotics, and the internet of things over 20 years (1998-2017). PloS one, 15(12), e0242984.

\bibitem{Kam2021b}
Ng, Kam KH, et al. "A systematic literature review on intelligent automation: Aligning concepts from theory, practice, and future perspectives." Advanced Engineering Informatics 47 (2021): 101246.

\bibitem{Hans2023}
Hans, Shivam, et al. "Artificial Intelligence with Robotics-A Bibliometric Study." 2023 10th International Conference on Computing for Sustainable Global Development (INDIACom). IEEE, 2023.

\bibitem{Riedl2019}
Riedl, M. O. (2019). Human‐centered artificial intelligence and machine learning. Human behavior and emerging technologies, 1(1), 33-36.

\bibitem{Shneiderman2020}
Shneiderman, B. (2020). Human-Centered Artificial Intelligence: Reliable, Safe \& Trustworthy. International Journal of Human–Computer Interaction, 36(6), 495–504.

\bibitem{Huang2022}
Huang, Wenlong, et al. "Inner monologue: Embodied reasoning through planning with language models." arXiv preprint arXiv:2207.05608 (2022).

\bibitem{IBMweb}
IBM design for AI, https://www.ibm.com/design/ai/

\bibitem{Doncieux2022}
Doncieux, S., Chatila, R., Straube, S. et al. Human-centered AI and robotics. AI Perspect 4, 1 (2022). 

\bibitem{He2022}
He, Hongmei, et al. "The challenges and opportunities of human-centered AI for trustworthy robots and autonomous systems." IEEE Transactions on Cognitive and Developmental Systems 14.4 (2021): 1398-1412.

\bibitem{Ball2017}
Ball, Rafael. An introduction to bibliometrics: New development and trends. Chandos Publishing, 2017.

\bibitem{Small1999}
H. Small, “Visualizing science by citation mapping,” Journal of the American Society for Information Science, vol. 50, no. 9, pp. 799–813, 1999.

\bibitem{Cobo2011}
Cobo, Manuel J., et al. "An approach for detecting, quantifying, and visualizing the evolution of a research field: A practical application to the Fuzzy Sets Theory field." Journal of informetrics 5.1 (2011): 146-166.

\bibitem{Cobo2019}
Moral-Muñoz, José A., et al. "Science mapping analysis software tools: A review." Springer handbook of science and technology indicators (2019): 159-185.

\bibitem{Cobo2012}
Cobo, Manuel J., et al. "SciMAT: A new science mapping analysis software tool." Journal of the American Society for information Science and Technology 63.8 (2012): 1609-1630.

\bibitem{VanEck2010}
van Eck, N.J., Waltman, L. Software survey: VOSviewer, a computer program for bibliometric mapping. Scientometrics 84, 523–538 (2010). 

\bibitem{Rane2024}
Rane, Nitin \& Mallick, Suraj \& Kaya, Ömer \& Rane, Jayesh. (2024). Emerging trends and future directions in machine learning and deep learning architectures.

\bibitem{Baruffaldi2020}
Baruffaldi, S., et al. (2020), "Identifying and measuring developments in artificial intelligence: Making the impossible possible", OECD Science, Technology and Industry Working Papers, No. 2020/05, OECD Publishing, Paris

\bibitem{Aggarwal2022}
Aggarwal, Karan, et al. "Has the future started? The current growth of artificial intelligence, machine learning, and deep learning." Iraqi Journal for Computer Science and Mathematics 3.1 (2022): 115-123.

\bibitem{Neal2017}
Roese, Neal J., and Kai Epstude. "The functional theory of counterfactual thinking: New evidence, new challenges, new insights." Advances in experimental social psychology. Vol. 56. Academic Press, 2017. 1-79.

\bibitem{White2004}
White, Steve R., et al. "An architectural approach to autonomic computing." International Conference on Autonomic Computing, 2004. Proceedings.. IEEE, 2004.

\end{thebibliography}
\end{document}